\documentclass[runningheads]{llncs}
\usepackage[T1]{fontenc}
\usepackage{amsmath}
\usepackage{graphicx}
\usepackage{amsfonts}
\usepackage{float}
\usepackage{booktabs}
\usepackage{hyperref}
\begin{document}
\title{Talk2SAM: Text-Guided Semantic Enhancement for Complex-Shaped Object Segmentation}
%
%
\titlerunning{Talk2SAM}
%
\author{Luka Vetoshkin\inst{1,2}\orcidID{0000-0002-0214-7917} \and
Dmitry Yudin\inst{1,3}\orcidID{0000-0002-1407-2633}} 
\authorrunning{L. Vetoshkin and D. Yudin}
%
\institute{
Moscow Institute of Physics and Technology (MIPT), Dolgoprudny 141701, Russia \and
Sber AI Lab, Moscow 117312, Russia \and
Artificial Intelligence Research Institute (AIRI), Moscow 117312, Russia
}
\maketitle              
\begin{abstract}
Segmenting objects with complex shapes, such as wires, bicycles, or structural grids, remains a significant challenge for current segmentation models, including the Segment Anything Model (SAM) and its high-quality variant SAM-HQ. These models often struggle with thin structures and fine boundaries, leading to poor segmentation quality. We propose \textbf{Talk2SAM}, a novel approach that integrates textual guidance to improve segmentation of such challenging objects. The method uses CLIP-based embeddings derived from user-provided text prompts to identify relevant semantic regions, which are then projected into the DINO feature space. These features serve as additional prompts for SAM-HQ, enhancing its ability to focus on the target object. Beyond improving segmentation accuracy, Talk2SAM allows user-controllable segmentation, enabling disambiguation of objects within a single bounding box based on textual input. We evaluate our approach on three benchmarks: BIG, ThinObject5K, and DIS5K. Talk2SAM consistently outperforms SAM-HQ, achieving up to +5.9\% IoU and +8.3\% boundary IoU improvements. Our results demonstrate that incorporating natural language guidance provides a flexible and effective means for precise object segmentation, particularly in cases where traditional prompt-based methods fail. The source code is available on GitHub: \href{https://github.com/richlukich/Talk2SAM}{\texttt{github.com/richlukich/Talk2SAM}}.

\keywords{Object Segmentation · Vision-Language Models · SAM · CLIP · Prompt Learning · Fine-Grained Structures}

\end{abstract}
\begin{figure}[t]
  \centering
  \includegraphics[width=0.9\textwidth]{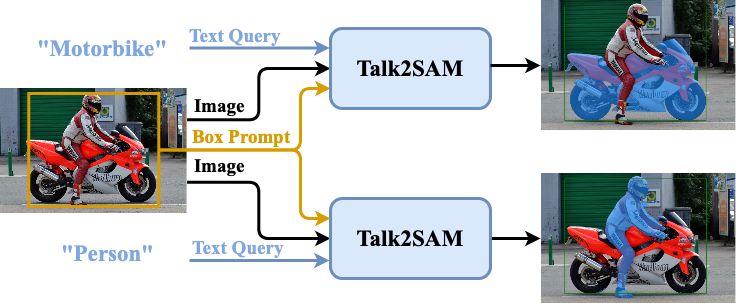}
  \caption{\textbf{Graphical Abstract.} The core idea of Talk2SAM is to guide open-world segmentation models using \textit{textual queries}, enabling precise selection of target objects. This approach overcomes the limitations of prompt-based methods such as SAM or HQ-SAM, where spatial prompts (e.g., boxes) may be insufficient to disambiguate complex object boundaries or distinguish between visually similar objects in close proximity.}
  \label{fig:ga}
\end{figure}

\section{Introduction}
Segmenting objects in natural images is a fundamental task in computer vision with widespread applications in robotics, medical imaging, autonomous driving, and image editing. Recent advances in foundation models, such as the Segment Anything Model (SAM)~\cite{kirillov2023segment}, have demonstrated remarkable generalization capabilities. However, these models often struggle with objects that possess complex boundaries or thin structures—such as wires, bicycles, and structural grids—where segmentation quality drops significantly.

To address these limitations, we introduce \textbf{Talk2SAM}, a novel method that integrates natural language understanding into the segmentation process. As illustrated in Figure~\ref{fig:ga}, our approach leverages CLIP~\cite{radford2021learning} embeddings derived from user-provided textual descriptions to guide segmentation toward semantically relevant regions. These embeddings are aligned with the DINO~\cite{caron2021emerging} feature space and injected as prompts into SAM-HQ~\cite{ke2023segment}, enabling more accurate and controllable segmentations. This language-driven control allows Talk2SAM to better distinguish between overlapping or visually similar objects, overcoming the limitations of traditional prompt types such as points or boxes.

Our key contributions are as follows:
\begin{itemize}
    \item We introduce \textbf{Talk2SAM}, a vision-language segmentation method that integrates CLIP, DINO, and SAM-HQ to segment complex-shaped objects by jointly leveraging semantic text prompts and high-resolution visual features.

    \item Unlike traditional prompt-based approaches, Talk2SAM incorporates language-based guidance to localize and disambiguate overlapping or structurally similar objects within a single bounding box.

    \item We conduct extensive experiments on three challenging datasets—BIG~\cite{cheng2020cascadepsp}, ThinObject5K~\cite{liew2021deep}, and DIS5K~\cite{qin2022highly}—demonstrating that Talk2SAM consistently outperforms SAM-HQ in both standard IoU and boundary IoU (bIoU) metrics.
\end{itemize}

\section{Related Work}

\textbf{SAM and its extensions.}The Segment Anything Model (SAM) introduced a prompt-based segmentation framework capable of generalizing across a wide range of objects using simple inputs such as points, boxes, or masks. While SAM demonstrates strong generality, it struggles with fine structures and objects requiring precise boundary delineation.

Several extensions have been proposed to address these limitations. SAM-HQ enhances boundary accuracy by refining high-frequency visual features, achieving consistently better mask quality than the original SAM. Other variants aim to improve semantic control or task adaptability: VRP-SAM~\cite{sun2024vrp} introduces visual reference prompts for contextual disambiguation; SAM-CLIP~\cite{wang2024sam} incorporates CLIP features to enable category-aware segmentation; and CAT-SAM~\cite{xiao2024cat} leverages adapter tuning for efficient task-specific customization.

These models illustrate the broader effort to enhance SAM’s spatial precision and semantic flexibility. Our approach, \textbf{Talk2SAM}, complements this line of work by introducing language-driven guidance for segmentation. In contrast to methods relying solely on visual prompting, Talk2SAM uses text prompts to inject semantic information, helping to distinguish structurally similar or overlapping objects—especially in cases involving thin or complex shapes.

Importantly, Talk2SAM is designed as a modular and model-agnostic enhancement: it can be integrated with any SAM-based variant to improve its performance without architectural modification. In this work, we adopt SAM-HQ as our primary baseline, as it significantly outperforms SAM in boundary-level accuracy. We further show that incorporating text-driven guidance into SAM-HQ yields additional improvements in both IoU and bIoU metrics.

\textbf{Vision-language segmentation with CLIP and DINO.}
Recent developments in vision-language models have enabled semantically informed segmentation through the use of textual prompts. A common approach involves leveraging CLIP to extract global or localized semantic features, which are then used to guide mask prediction. CLIPSeg~\cite{luddecke2022image} was among the first to demonstrate this idea, conditioning a segmentation decoder on CLIP-derived embeddings from image and text inputs. While effective in zero-shot scenarios, CLIPSeg suffers from limited spatial precision, particularly in cluttered or fine-grained regions.

To improve spatial grounding, DenseCLIP~\cite{rao2022denseclip} introduces context-aware prompts and dense supervision, improving alignment between language and visual features. Similarly, LSeg~\cite{li2022language} aligns a supervised segmentation backbone with CLIP’s latent space, enabling open-vocabulary prediction, though with limited control over instance-level segmentation. X-Decoder~\cite{zou2023generalized} extends this paradigm by unifying decoding across image and text modalities for a variety of dense prediction tasks, but prioritizes generality over precise structure-aware segmentation.

A more recent and closely related method is \textbf{Talk2DINO}~\cite{barsellotti2024talking}, which bridges CLIP and DINO by projecting text-derived semantics into the high-resolution DINO feature space. This enables semantically controllable detection and forms the foundation for multimodal prompt-based reasoning at fine spatial granularity. We adopt Talk2DINO in our work as a semantic prior to compute dense similarity maps from text prompts. By injecting these maps into SAM-HQ, \textbf{Talk2SAM} extends semantic projection into the segmentation domain, enabling accurate and controllable segmentation of thin and structurally complex objects guided purely by natural language.

\section{Method}

In this section, we present \textbf{Talk2SAM}, a vision-language segmentation method that enhances object segmentation by incorporating textual semantics into the high-quality SAM model (SAM-HQ). The method leverages language as an additional modality to provide semantic guidance for segmentation, particularly useful for fine-structured or ambiguous objects.

\begin{figure}[t]
    \centering
    \includegraphics[width=\linewidth]{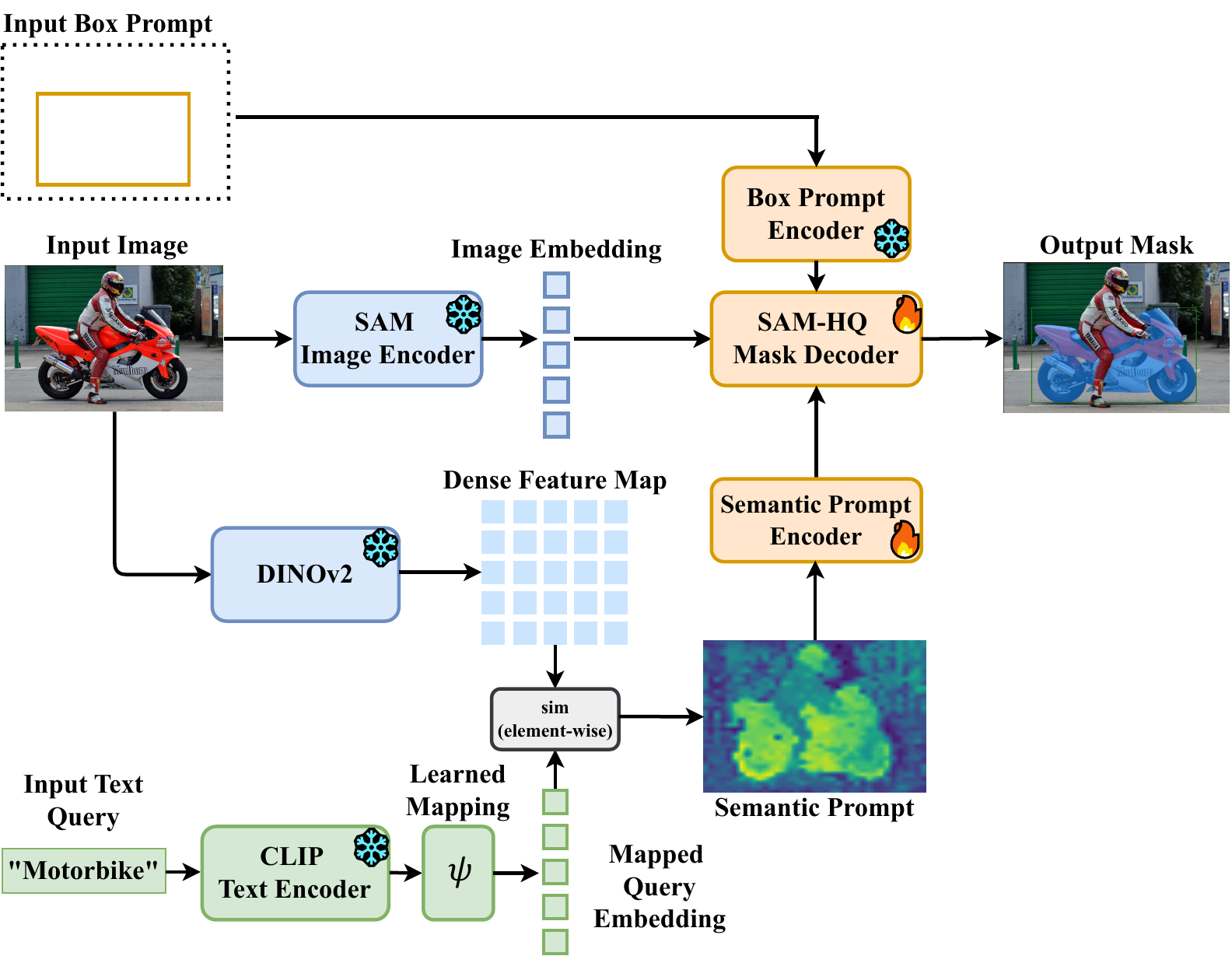}
    \caption{Overview of the Talk2SAM framework. Given an input image and a user-defined text prompt, CLIP extracts a global text embedding and computes a relevance map with the image. This relevance is projected into DINO’s visual space to produce a dense semantic prompt, which is injected into SAM-HQ's decoder for refined segmentation.}
    \label{fig:pipeline}
\end{figure}

Figure~\ref{fig:pipeline} illustrates the overall pipeline. Talk2SAM consists of three main stages:

\begin{enumerate}
    \item \textbf{Text-to-Semantic Mapping:} The user provides a textual object category (e.g., \textit{"wire"} or \textit{"bicycle"}), rather than a full natural language description. This category is encoded using CLIP’s text encoder to produce a semantic embedding. Following Talk2DINO, we apply a nonlinear learned mapping to project this embedding into the visual space of DINOv2 features. Specifically, the projection $\psi(\cdot)$ is composed of two affine transformations with a hyperbolic tangent activation:
    \begin{equation}
    \psi(t) = W_b^\top \left( \tanh\left( W_a^\top t + b_a \right) \right) + b_b,
    \end{equation}
    where $W_a$, $W_b$ are learnable projection matrices, and $b_a$, $b_b$ are bias terms. This warping function adapts the CLIP text embeddings to the visual patch embedding space used by DINOv2, ensuring better alignment between modalities.

    \item \textbf{Projection to DINO Space:} Following the approach in Talk2DINO, a nonlinear projection function $\psi(\cdot)$ is used to align CLIP text embeddings with the DINOv2 visual feature space. This alignment is guided by the internal attention mechanism of DINOv2, which naturally segments the image into semantic regions.

    In the final transformer layer of DINOv2, $N$ attention maps $A_i \in \mathbb{R}^{H/P \times W/P}$ are computed between the [CLS] token and spatial patches. Each attention map $A_i$ corresponds to one attention head and highlights a different semantic region.
    
    For each attention map, a visual embedding $v_{A_i} \in \mathbb{R}^{D_v}$ is obtained as a weighted average of the patch features $v[h, w]$ using the corresponding attention weights:
    \begin{equation}
    v_{A_i} = \sum_{h,w} \mathrm{softmax}(A_i)[h,w] \cdot v[h,w].
    \end{equation}
    
    The similarity between each $v_{A_i}$ and the projected text embedding $\psi(t)$ is then computed using cosine similarity:
    \begin{equation}
    \mathrm{sim}(v_{A_i}, t) = \frac{v_{A_i} \cdot \psi(t)^\top}{\|v_{A_i}\| \cdot \|\psi(t)\|}.
    \end{equation}
    
    To select the most relevant alignment, the maximum similarity score across all heads is taken:
    \begin{equation}
    s(t) = \max_{i=1,\ldots,N} \mathrm{sim}(v_{A_i}, t).
    \end{equation}
    
    This procedure promotes alignment between the text embedding and the most semantically relevant visual region, enabling the generation of dense features that capture both spatial and linguistic intent.

    \item \textbf{Semantic Feature Conditioning:} The similarity map, computed from the alignment between the projected text embedding and DINOv2 attention-based visual features, is used to guide the segmentation process. This map is encoded via a lightweight convolutional module structurally analogous to the mask input encoder used in the original SAM-HQ. However, unlike SAM-HQ, where the mask embedding is fixed and task-agnostic, our encoder is \emph{trainable} and optimized jointly with the rest of the model. This adaptation allows it to better capture the semantics of the input prompt. The resulting mask embedding is then fed into the SAM-HQ decoder alongside geometric prompts, enabling joint reasoning over both spatial structure and language-derived semantic intent.

\end{enumerate}

\subsection{Training Details}

During training, we freeze the parameters of the DINOv2 image encoder and the SAM encoder (ViT backbone) in all settings. The training strategy for CLIP varies depending on data availability: in low-data regimes, the entire CLIP model (both vision and text encoders) remains frozen to avoid overfitting; in large-scale settings, we fine-tune only the final projection layer of the CLIP text encoder to better adapt to task-specific semantics.

We adopt the CLIP ViT-B/16 and DINOv2-ViT-L/14 backbones, as in Talk2DINO, to compute a similarity map between text and visual features. The resulting similarity maps are of size $37 \times 37$ and are bilinearly upsampled to $256 \times 256$ before being fed into the segmentation pipeline. These maps serve as a learned semantic prior and are treated as input to the mask prompt encoder module from the original SAM-HQ architecture.

Notably, this encoder was originally designed to consume binary mask inputs, and thus remains incompatible with text-derived similarity maps in its fixed form. In our method, we retrain this module to adapt to the continuous, semantically dense structure of the similarity map. This modification enables the decoder to process semantic information in a format it was not originally designed to handle.

The following components are updated during training:
\begin{itemize}
    \item A learned nonlinear mapping that transforms CLIP text embeddings into the DINOv2-aligned similarity space.
    \item The SAM-HQ decoder, responsible for generating high-resolution segmentation masks.
    \item The prompt encoder that processes the similarity map into a mask embedding; unlike in SAM-HQ, this module is fully trainable.
\end{itemize}

We train the model using stochastic gradient descent (SGD) with a learning rate of 0.001 and a batch size of 4, for 30 epochs on a single NVIDIA Tesla V100 GPU. We employ the standard SAM-HQ loss formulation, which combines binary cross-entropy (BCE) loss and dice loss to supervise the predicted masks.

To quantify the efficiency of our training regime, we report the number of trainable and total parameters under different configurations in Table~\ref{tab:trainable_params}. Despite integrating large-scale vision-language backbones such as CLIP and DINOv2, our method maintains a small training footprint: less than 1\% of parameters are updated during optimization. This highlights the practicality of Talk2SAM, even in resource-constrained scenarios.

Despite incorporating large-scale vision-language backbones, Talk2SAM remains memory-efficient. When trained with a batch size of 1 on a single Tesla V100 GPU, peak memory usage reaches approximately 10.7\,GB, compared to 10.1\,GB for SAM-HQ. This marginal increase demonstrates that our semantic guidance mechanism introduces minimal overhead, preserving the practicality of training under standard hardware constraints.

\begin{table}[t]
\centering
\caption{Number of parameters in SAM-HQ and Talk2SAM under different training regimes.}
\begin{tabular}{lccc}
\toprule
\textbf{Model} & \textbf{Trainable Params} & \textbf{Total Params} & \textbf{Trainable (\%)} \\
\midrule
SAM-HQ & 5.65M & 641M & 0.88\% \\
Talk2SAM (CLIP frozen) & 7.22M & 1.096B & 0.66\% \\
Talk2SAM (CLIP partial) & 10.64M & 1.096B & 0.97\% \\
\bottomrule
\end{tabular}
\label{tab:trainable_params}
\end{table}

\section{Experiments}

\subsection{Datasets}

We evaluate our method on three publicly available datasets, each focusing on fine-grained or complex object structures where segmentation is particularly challenging. In all datasets, text prompts were automatically extracted from the filenames, which typically reflect the dominant object or region depicted in the image.

\textbf{ThinObject5K} is a benchmark designed for evaluating segmentation of thin and elongated structures such as wires, ropes, and fences. It contains 5,000 images with high-quality annotations that emphasize fine boundaries and delicate shapes. Textual categories were extracted directly from image filenames, which consistently describe the object class of interest.

\textbf{DIS5K} (Detail-Injection Segmentation) includes 5,470 high-resolution images with pixel-accurate masks focusing on boundary precision. Unlike the other datasets, DIS5K contains filenames that occasionally describe scenes or actions (e.g., “skiing”) rather than concrete objects. To ensure that the text prompts aligned semantically with visible objects, we manually filtered out samples with ambiguous or non-informative file names. After filtering, we retained 2,777 images for training and 457 for validation.

\textbf{BIG} (Boundary-aware Instance Grouping) offers class-agnostic, high-resolution masks tailored for evaluating general-purpose segmentation. The dataset emphasizes complex layouts and fine object contours. As with ThinObject5K, we used filenames to extract category labels that serve as text prompts during inference.

These datasets were selected primarily for two reasons. First, their file naming conventions allow for the reliable extraction of textual object categories, which is essential for evaluating our text-driven segmentation method. Second, all three datasets provide high-quality, pixel-accurate masks with detailed boundary annotations. This enables precise quantitative comparison with the baseline SAM-HQ model, particularly in scenarios involving thin structures, fine contours, and complex object layouts.

\begin{figure}[t]
\centering
\includegraphics[width=\linewidth]{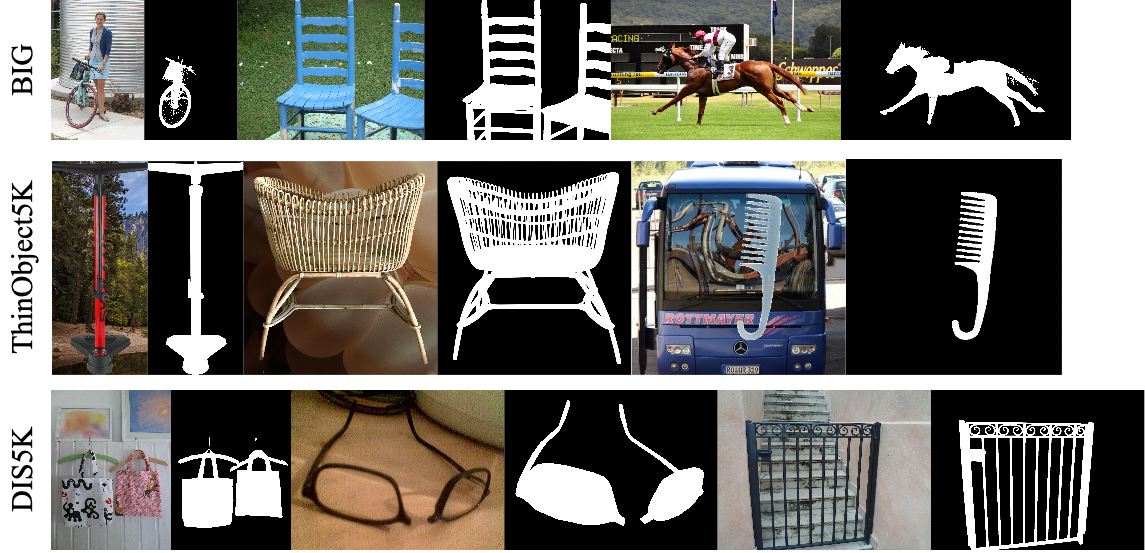}
\caption{
Representative image and mask pairs from the datasets used in our study. Although not shown visually, textual categories used as prompts in our method were automatically extracted from the filenames of these samples. The figure highlights the diversity and structural complexity of the segmented objects, motivating the use of text-guided segmentation.
}
\label{fig:dataset_examples}
\end{figure}

To illustrate the nature of our data sources, Figure~\ref{fig:dataset_examples} presents examples of original images and their corresponding ground-truth masks from the datasets used in our study. While text prompts are not visualized, they were automatically extracted from the filenames associated with these images.

\subsection{Evaluation Metrics}
We evaluate segmentation quality using two standard metrics: mean Intersection-over-Union (mIoU) and mean boundary Intersection-over-Union (mBIoU).

\textbf{mIoU} is the average intersection-over-union between predicted and ground truth masks across all test samples. It measures overall segmentation accuracy by comparing pixel-level overlap.

In addition to standard mean Intersection-over-Union (mIoU), we report results using \textbf{mean Boundary IoU (mBIoU)}~\cite{cheng2021boundary}, a boundary-aware extension of mIoU that emphasizes the quality of object contours. Unlike mIoU, which treats all pixels equally, mBIoU increases the evaluation weight of pixels near object boundaries, making it particularly sensitive to fine structural details. This is especially important for datasets such as ThinObject5K and DIS5K, where slight misalignments at object edges may lead to significant perceptual degradation. By including mBIoU, we aim to capture the segmentation accuracy more faithfully in scenarios involving thin, elongated, or complex-shaped objects.

Both metrics are reported as averages across the entire dataset. In all experiments, we compare our method (Talk2SAM) with the baseline SAM-HQ using the same image encoders for fair comparison.

\begin{table}[H]
\centering
\caption{Results on the ThinObject5K dataset: segmentation performance and inference time (ms) using different ViT backbones.}
\label{tab:thinobject5k}
\begin{tabular}{lccc}
\hline
\textbf{Method} & \textbf{mIoU} & \textbf{mBIoU} & \textbf{Time (ms)} \\
\hline
SAM ViT-H        & 0.686      & 0.591      & \textbf{557} \\
SAM-HQ ViT-H     & 0.870      & 0.759      & 560 \\
Talk2SAM ViT-H   & \textbf{0.929} & \textbf{0.842} & 720 \\
\hline
SAM-HQ ViT-B     & 0.811      & 0.695      & \textbf{142} \\
Talk2SAM ViT-B   & \textbf{0.921}      & \textbf{0.830}      & 302 \\
\hline
SAM-HQ ViT-L     & 0.853      & 0.737      & \textbf{329} \\
Talk2SAM ViT-L   & \textbf{0.926}      & \textbf{0.839}      & 489 \\
\hline
\end{tabular}
\end{table}

\begin{table}[H]
\centering
\caption{Results on BIG dataset using ViT-H encoder.}
\label{tab:big}
\begin{tabular}{lcc}
\hline
\textbf{Method} & \textbf{mIoU} & \textbf{mBIoU} \\
\hline
SAM ViT-H        & 0.902      & 0.784 \\
SAM-HQ ViT-H     & 0.929      & 0.824      \\
Talk2SAM ViT-H   & \textbf{0.947}      & \textbf{0.866 }     \\
\hline
\end{tabular}
\end{table}

\begin{table}[H]
\centering
\caption{Results on DIS5K dataset using ViT-H encoder.}
\label{tab:dis5k}
\begin{tabular}{lcc}
\hline
\textbf{Method} & \textbf{mIoU} & \textbf{mBIoU} \\
\hline
SAM ViT-H        & 0.573     & 0.497  \\
SAM-HQ ViT-H     &  0.723     &  0.660     \\
Talk2SAM ViT-H   &  \textbf{0.749}     &  \textbf{0.663 }    \\
\hline
\end{tabular}
\end{table}

\section{Qualitative Results}

\begin{figure}[htbp]
\centering
\includegraphics[width=\linewidth]{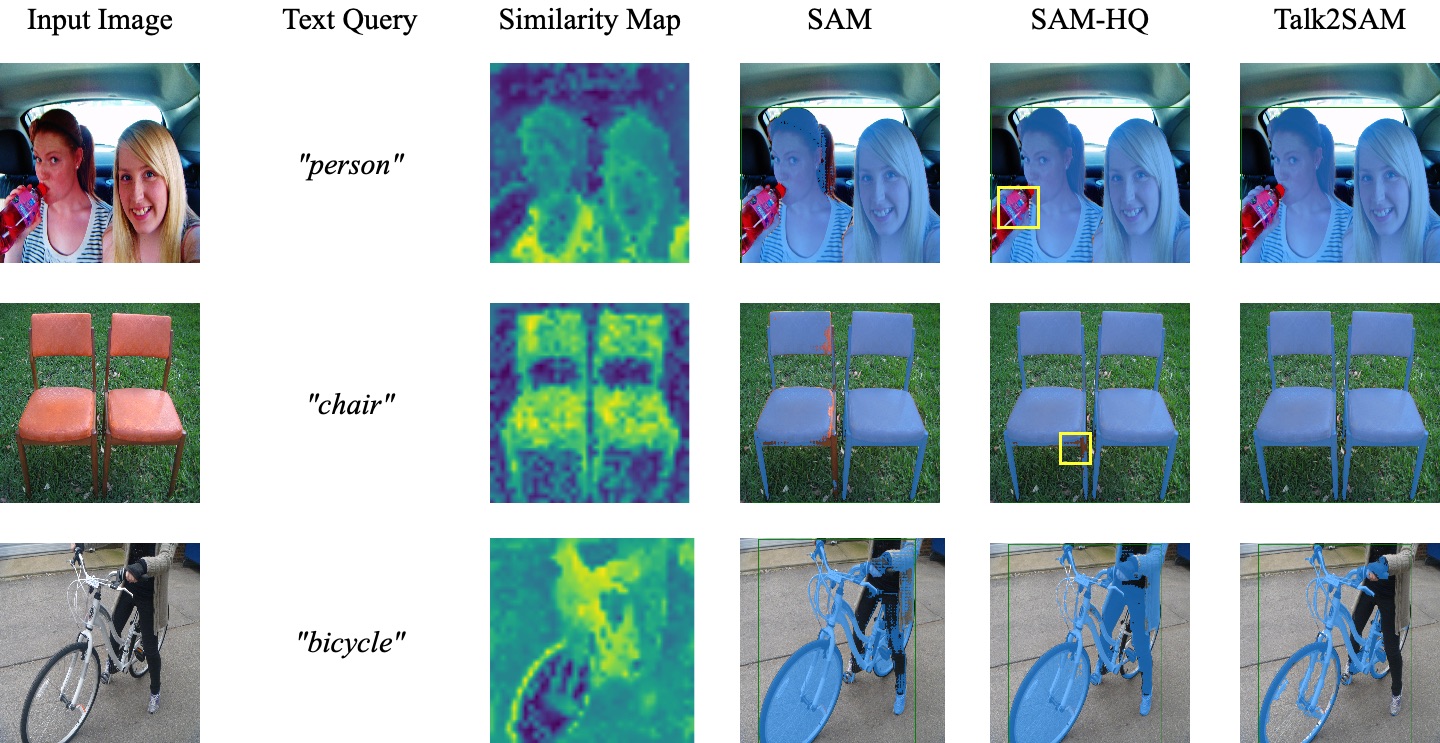}
\caption{
Qualitative results on the BIG dataset. Each row shows: the input image, the text query, the similarity map from CLIP-DINOv2 features, and the predicted masks from SAM, SAM-HQ, and Talk2SAM. While SAM and SAM-HQ often produce incomplete or imprecise masks, Talk2SAM leverages textual guidance to accurately segment target objects, even in complex scenes.
}
\label{fig:qual_big}
\end{figure}

Figure~\ref{fig:qual_big} presents qualitative comparisons on the BIG dataset. We showcase the performance of SAM, SAM-HQ, and our method, Talk2SAM, across diverse examples using text queries such as \textit{“person”}, \textit{“chair”}, and \textit{“bicycle”}. Each example includes the input image, the text query, the similarity map generated from projected CLIP-DINOv2 features, and the segmentation results from each method.

As illustrated, SAM frequently exhibits issues such as ambiguous object boundaries or partial segmentations. SAM-HQ improves upon these by refining object edges, but it often struggles with contextual disambiguation. In contrast, Talk2SAM effectively integrates textual semantics to localize and segment the intended objects more accurately. This is particularly evident in the bicycle example, where Talk2SAM cleanly separates the bike from the background and surrounding clutter.

\begin{figure}[htbp]
\centering
\includegraphics[width=\linewidth]{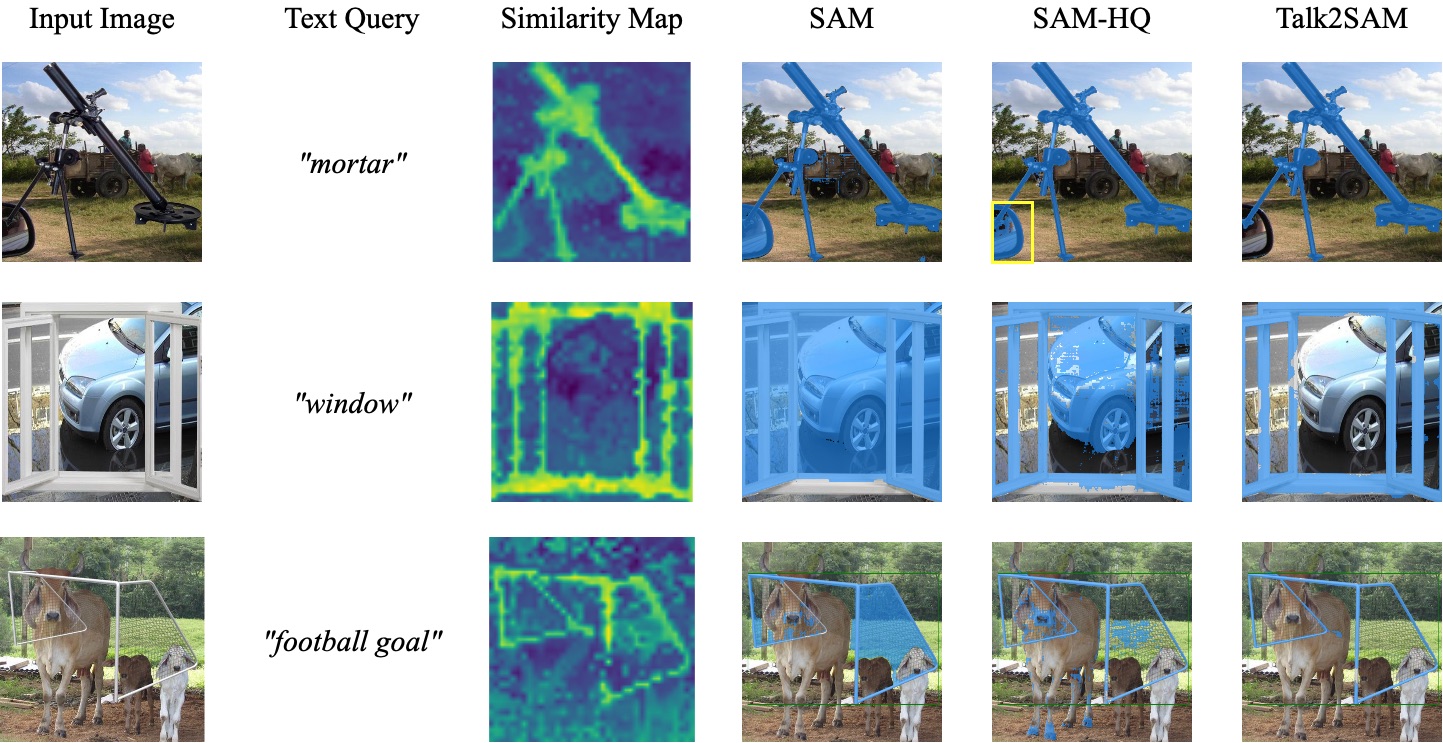}
\caption{
Qualitative results on the ThinObject5K dataset. Each row includes: the input image, the corresponding text query, the similarity map obtained from projected CLIP-DINOv2 features, and segmentation masks from SAM, SAM-HQ, and Talk2SAM. Talk2SAM outperforms baselines in preserving continuity and shape accuracy, particularly in fine-grained and thin-object scenarios.
}
\label{fig:qual_thin}
\end{figure}

Figure~\ref{fig:qual_thin} highlights qualitative results on the ThinObject5K dataset, which focuses on challenging thin and elongated structures. In such cases, even small segmentation errors can significantly degrade perceived quality. Talk2SAM consistently captures the full extent of these delicate structures—such as wires, railings, and goalposts—where SAM and SAM-HQ often fail to preserve continuity or context. This demonstrates the advantage of incorporating text-driven semantic priors for structure-aware segmentation.

\begin{figure}[htbp]
\centering
\includegraphics[width=\linewidth]{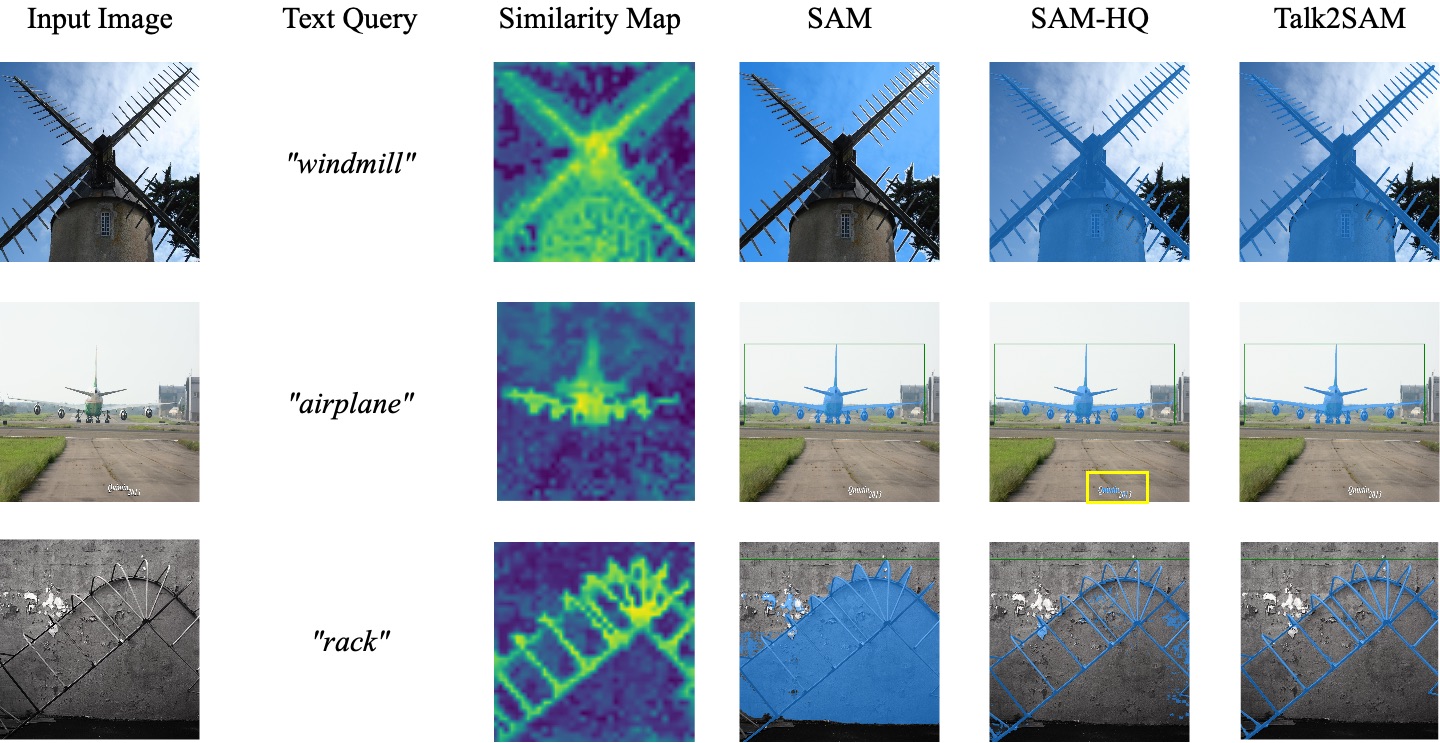}
\caption{
Qualitative results on the DIS5K dataset. Each row displays: the input image, the corresponding text query, the similarity map, and the segmentations by SAM, SAM-HQ, and Talk2SAM. Talk2SAM maintains high segmentation fidelity in scenes with complex structure and fine detail.
}
\label{fig:qual_dis5k}
\end{figure}

Figure~\ref{fig:qual_dis5k} presents additional examples from the DIS5K benchmark, emphasizing Talk2SAM’s robustness on detailed and visually complex scenes. Our method demonstrates superior capability in segmenting both small and intricate regions, aided by the semantic cues provided by the text prompts.

\begin{figure}[!h]
\centering
\includegraphics[width=\linewidth]{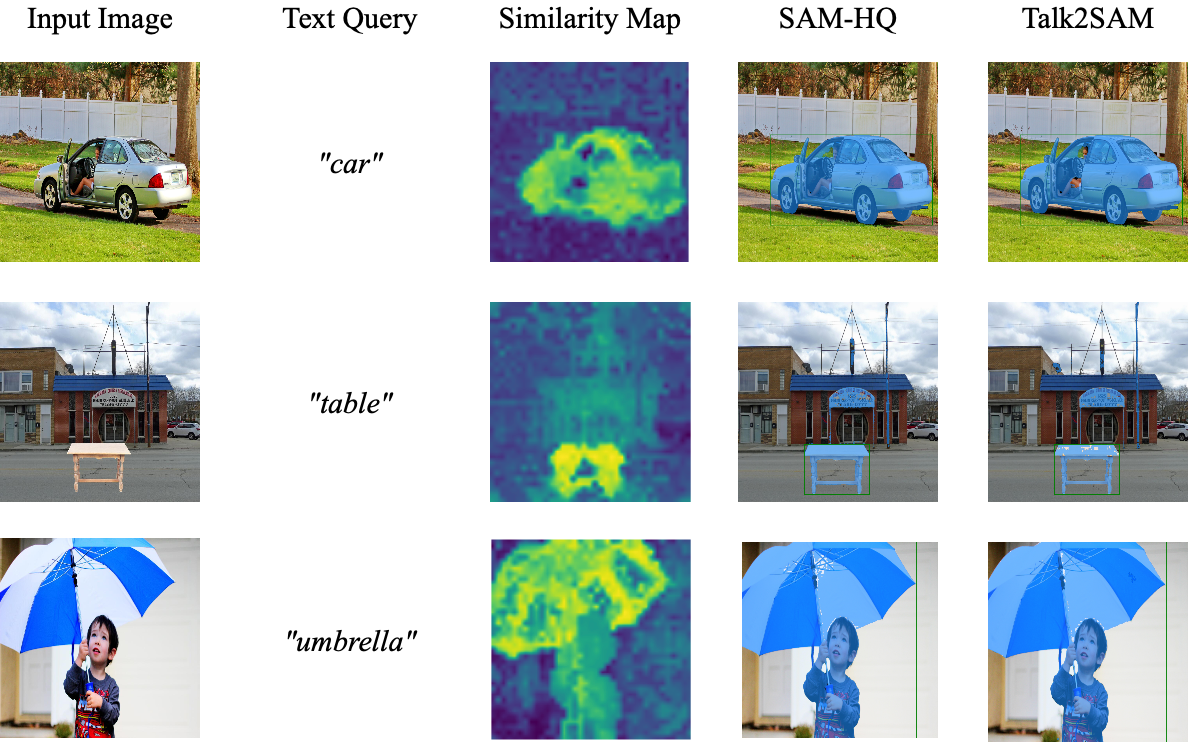}
\caption{
Challenging cases where similarity maps align well with the text prompts but segmentation performance suffers. Each row shows: the input image, text query, similarity map from CLIP-DINOv2, and predicted masks from SAM-HQ and Talk2SAM. These examples highlight that accurate semantic localization alone is not sufficient—segmentation models like SAM-HQ may still fail when the visual features deviate from commonly seen patterns. Talk2SAM mitigates this by aligning semantic intent with spatial prediction.
}
\label{fig:qual_bad}
\end{figure}

Figure~\ref{fig:qual_bad} illustrates several failure cases where the similarity maps accurately reflect the text query, yet the final segmentation results are suboptimal. These examples reveal an important limitation: although the similarity map derived from CLIP-DINOv2 features effectively localizes the semantically relevant regions (e.g., the shape of the umbrella or the silhouette of the car), the downstream segmentation quality is still constrained by the inherent limitations of the SAM and SAM-HQ models.

Notably, SAM-HQ often inherits the weaknesses of its base model and struggles to segment less common or partially occluded objects. Since SAM-HQ is not explicitly aware of the semantic intent behind the similarity map, its refinement step cannot fully compensate for failures in understanding the object of interest. In contrast, Talk2SAM leverages textual priors to bridge this gap, allowing it to produce more coherent and context-aware segmentations, even when dealing with ambiguous or underrepresented categories.

\section{Conclusion}

In this work, we introduced \textbf{Talk2SAM}, a novel method for segmenting complex-shaped objects—such as wires, grids, and bicycle frames—using natural language prompts. Our approach enhances the segmentation performance of SAM-HQ by injecting semantic guidance from CLIP into the DINO feature space, enabling more precise localization and differentiation of fine structures that traditional prompt-based models struggle to resolve. Talk2SAM further supports user-controllable segmentation and disambiguation within a single bounding box, offering a flexible and interpretable interface for complex visual scenes.

An important aspect of our method is its \emph{modularity and generality}. Talk2SAM does not require architectural changes to the underlying segmentation model and can be seamlessly integrated with any SAM-based variant. As such, it has the potential to improve the performance of a wide range of promptable segmentation models by incorporating semantic-level reasoning through language.

Despite its advantages, the method has several limitations. It relies on CLIP embeddings, which may struggle to represent rare or domain-specific concepts. While the DINO projection enhances spatial alignment, final mask quality remains influenced by the resolution and capacity of the backbone decoder. Additionally, Talk2SAM assumes that the target object can be meaningfully described using a short textual prompt, which may not generalize to all use cases.

For future work, we plan to explore bidirectional refinement between vision and language, allowing prompts to adapt dynamically to scene context. We also aim to extend Talk2SAM to multi-object and referring segmentation, and to incorporate language models capable of generating textual queries automatically from images. Evaluating Talk2SAM in real-world domains such as medical imaging or industrial inspection will be critical for validating its practical utility.

\bibliographystyle{splncs04}
\bibliography{references}

\end{document}